\documentclass[sigconf]{acmart}
\AtBeginDocument{%
  }

\usepackage{graphics}
\usepackage{graphicx}
\usepackage{xspace}
\usepackage{subcaption}
\usepackage{amsmath}
\usepackage{amsthm}
\usepackage{mathrsfs}
\usepackage{array}
\usepackage{textcomp}
\usepackage{xcolor}
\usepackage{url}
\usepackage{diagbox}
\usepackage{color}
\usepackage{booktabs}
\usepackage{listings}
\usepackage{tcolorbox}
\usepackage{setspace}
\usepackage{multirow}
\usepackage{booktabs}    
\usepackage{tabularx}    
\usepackage{array}       
\usepackage{makecell}    
\usepackage{enumitem}
\usepackage{balance}

\newcommand{\ts}[1]{\textsuperscript{\textcolor{black}{#1}}}

\newcommand{\myparagraph}[1]{\vspace{0.5mm} \noindent \textbf{#1}.}

\copyrightyear{2026}
\acmYear{2026}
\setcopyright{cc}
\setcctype{by}
\acmConference[WWW '26] {Proceedings of the ACM Web Conference 2026}{April 13--17, 2026}{Dubai, United Arab Emirates.}
\acmBooktitle{Proceedings of the ACM Web Conference 2026 (WWW '26), April 13--17, 2026, Dubai, United Arab Emirates}
\acmISBN{979-8-4007-2307-0/2026/04}
\acmDOI{10.1145/3774904.3792389}

\begin{document}

\title[Towards Open-World Retrieval-Augmented Generation on Knowledge Graph]{Towards Open-World Retrieval-Augmented Generation on Knowledge Graph: A Multi-Agent Collaboration Framework}


\author{Jiasheng Xu}
\orcid{0009-0000-9118-8288}
\affiliation{%
  \department{School of Computer Science and Cyber Engineering,}
  \institution{Guangzhou University,}
  \city{Guangzhou}
  \state{Guangdong}
  \country{China}}
\affiliation{%
  \institution{Institute of Automation, Chinese Academy of Sciences,}
  \city{Beijing}
  \country{China}}
\email{jiasheng@e.gzhu.edu.cn}
\authornote{Equal contribution.}

\author{Mingda Li}
\orcid{0000-0002-7240-2063}
\affiliation{%
  \institution{Institute of Automation, Chinese Academy of Sciences,}
  \city{Beijing}
  \country{China}}
\email{limingda2018@ia.ac.cn}
\authornotemark[1]

\author{Yongqiang Tang}
\orcid{0000-0001-9333-8200}
\affiliation{%
  \institution{Institute of Automation, Chinese Academy of Sciences,}
  \city{Beijing}
  \country{China}}
\email{yongqiang.tang@ia.ac.cn}
\authornote{Corresponding author.}

\author{Peijie Wang}
\orcid{0009-0005-1584-0058}
\affiliation{%
  \institution{Institute of Automation, Chinese Academy of Sciences,}
  \city{Beijing}
  \country{China}}
\email{wangpeijie2023@ia.ac.cn}

\author{Wensheng Zhang}
\orcid{0000-0003-0752-941X}
\affiliation{%
  \department{School of Computer Science and Cyber Engineering,}
  \institution{Guangzhou University,}
  \city{Guangzhou}
  \state{Guangdong}
  \country{China}}
\affiliation{%
  \institution{Institute of Automation, Chinese Academy of Sciences,}
  \city{Beijing}
  \country{China}}
\email{zhangwenshengia@hotmail.com}


\begin{abstract}
Large Language Models (LLMs) have demonstrated strong capabilities in web search and reasoning. However, their dependence on static training corpora makes them prone to factual errors and knowledge gaps. Retrieval-Augmented Generation (RAG) addresses this limitation by incorporating external knowledge sources, especially structured Knowledge Graphs (KGs), which provide explicit semantics and efficient retrieval. Existing KG-based RAG approaches, however, generally assume that anchor entities are accessible to initiate graph traversal, which limits their robustness in open-world settings where accurate linking between the user query and the KG entity is unreliable. To overcome this limitation, we propose \textbf{AnchorRAG}, a novel multi-agent collaboration framework for open-world RAG without the predefined anchor entities. Specifically, a predictor agent dynamically identifies candidate anchor entities by aligning user query terms with KG nodes and initializes independent retriever agents to conduct parallel multi-hop explorations from each candidate. Then a supervisor agent formulates the iterative retrieval strategy for these retriever agents and synthesizes the resulting knowledge paths to generate the final answer. This multi-agent collaboration framework improves retrieval robustness and mitigates the impact of ambiguous or erroneous anchors. Extensive experiments on four public benchmarks demonstrate that AnchorRAG significantly outperforms existing baselines and establishes new state-of-the-art results on the real-world reasoning tasks. The datasets along with our code are available at \href{https://github.com/Jayson3831/AnchorRAG}{https://github.com/Jayson3831/AnchorRAG}.
\end{abstract}





\maketitle

\newcommand\webconfavailabilityurl{https://doi.org/10.5281/zenodo.18310915}
\ifdefempty{\webconfavailabilityurl}{}{
\begingroup\small\noindent\raggedright\textbf{Resource Availability:}\\
The source code of this paper has been made publicly available at \url{\webconfavailabilityurl}.
\endgroup
}

\section{Introduction}

Large Language Models (LLMs)~\cite{achiam2023gpt, yang2025qwen3} are typically defined as deep learning models with a massive number of parameters, trained on large-scale corpora in a self-supervised manner. Their internal parameterization allows for an implicit representation of external knowledge. During the inference stage, the chain-of-Thought (CoT)~\cite{wei2022chain} method guides the models to ``think step-by-step" through carefully designed prompts. It helps the models better handle logically complex or multi-step reasoning tasks, achieving a better performance in natural language processing tasks such as web search~\cite{li2025towards}. Despite these advances, LLMs are fundamentally limited by the incompleteness and inaccuracies in their static training corpora. This often results in factual hallucinations~\cite{huang2025survey}, especially in knowledge-intensive tasks, significantly hindering their reliability and deployment in real-world applications.

To address these limitations, recent studies~\cite{jeong2024adaptive, xia2025improving} have investigated augmenting LLMs with external knowledge sources to improve factual consistency and reasoning. While fine-tuning \cite{hu2022lora} LLM with the external knowledge is computationally expensive and inflexible, a promising solution is Retrieval-Augmented Generation (RAG)~\cite{lewis2020retrieval}, which combines real-time retrieval from external sources with generative modeling. To improve the retrieval efficiency of the external sources, some researchers apply the knowledge graph (KG) to represent the external information as the structured triples, i.e., (entity, relation, entity). Then the core issue of RAG on the knowledge graph is to locate the relevant triples for enhancing the reasoning capability.

Among existing solutions, iterative retrieval methods~\cite{sun2023think, ma2025debate, chen2024plan, xu2024generate, wang2025reasoning} has become a research hotspot in the field, due to the fact that they can handle the complex user queries by dynamically constructing and refining explainable reasoning paths through multi-round ``retrieval-generation". Typically, these RAG methods can be divided in two steps: (1) identifying a query-aware anchor entity to initiate retrieval; (2) performing iterative retrieval to construct the reasoning paths. Through the above steps, these methods can effectively exploit the external knowledge related to the query and improve the performance in the downstream reasoning task.

\begin{figure}[t]
	\centering
	\includegraphics[width=\linewidth]{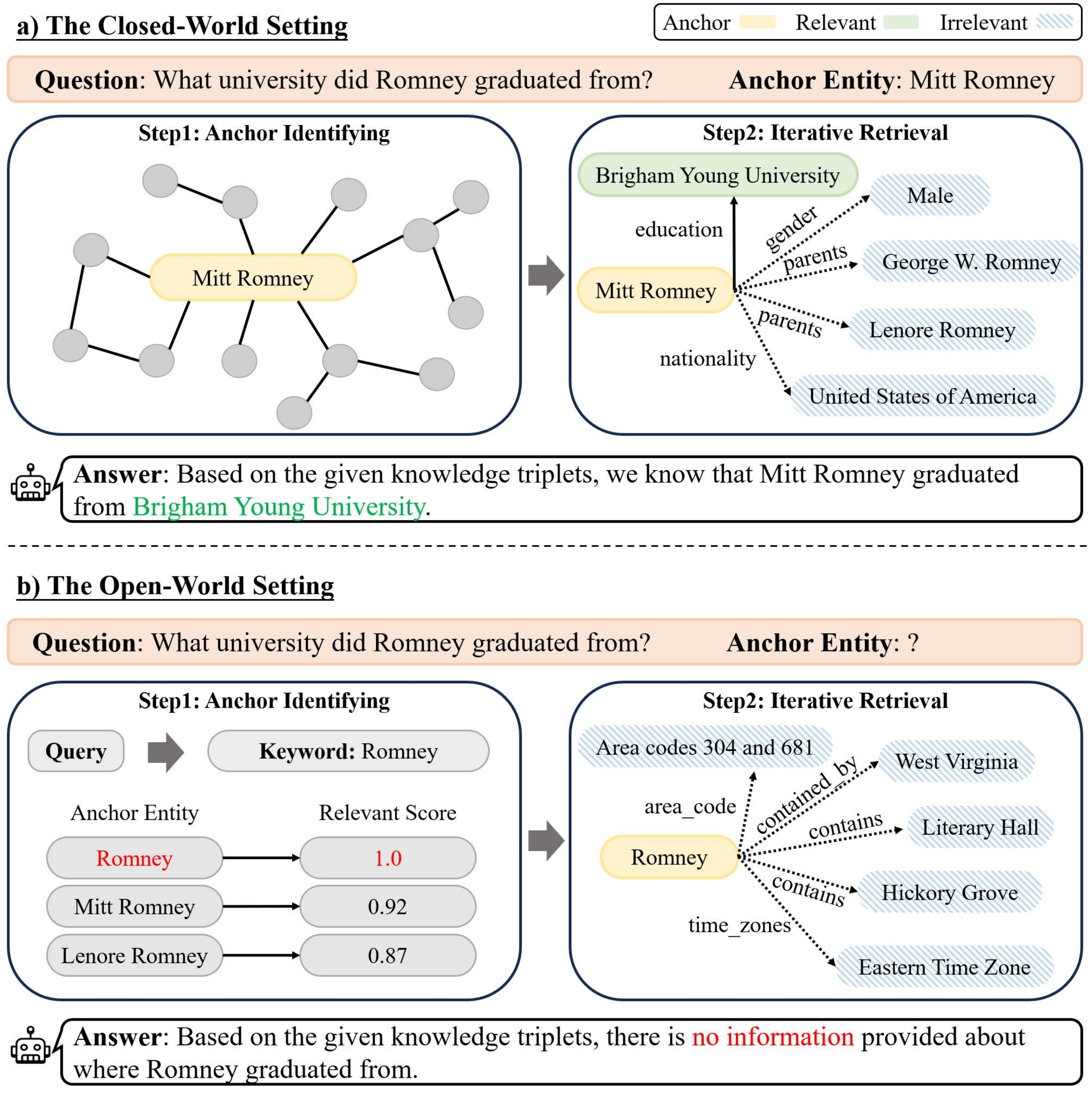}
	\caption{Illustration of the difference between the closed-world and the open-world setting in RAG. }
	\label{fig:comp-case}
\end{figure}

However, existing KG-based RAG methods almost follow the closed-world setting where the anchor entity is accessible and exists in the given KG. As shown in the Figure \ref{fig:comp-case} (a), the anchor entity ``Mitt Romney" is pre-defined for the question ``What university did Romney graduated from?". The RAG methods can directly locate the target entity for the following retrieval. Actually, user questions usually conform to the open-world setting where the anchor entity is unavailable. In that case, existing methods usually borrow from entity linking models and extract the keywords in the query to identify the candidate anchor entity by semantic matching. Due to name abbreviations and aliases, these methods usually suffer from the issue caused by imprecise or partial matching as shown in Figure \ref{fig:comp-case}(b). Then the challenge of the open-world RAG lies in how to accurately identify anchor entities to retrieve the relevant information.


To tackle this issue, we propose \textbf{AnchorRAG}, a multi-agent collaborative (MAC) framework that can perform RAG without the predefined anchor entities, thereby enabling more effective knowledge retrieval to enhance the LLM reasoning performance. Specifically, our framework follows a pipeline-based MAC framework, and a predictor agent first extracts the keywords from the given question and applies a semantic match model to obtain the candidate entities by their name description and structure neighborhood. Then, multiple retriever agents initiate parallel graph traversal with each candidate, conducting iterative retrieval to obtain the pivotal knowledge for the following reasoning. Finally, a supervisor agent will synthesize the retrieved evidence and determine whether an answer can be generated and formulate the retrieval process. The main contributions of this paper are summarized as follows:

\begin{itemize}
    \item \textbf{General Aspect.} We emphasize the importance of identifying the accurate anchor entity for the open-world RAG. By integrating multi-agent collaboration into the workflow of question answering, we can enhance the generality of RAG methods and improve the performance in the open-world setting.

    \item \textbf{Methodologies.} For anchor identifying, we design an entity grounding strategy to locate the candidate anchors by entity name description and structural neighborhood. For knowledge retrieval, we propose a novel retrieval method with candidate ranking and evidence validation, which can capture the resultful knowledge paths to boost the LLM reasoning.
    
    \item \textbf{Experimental Findings.} Extensive experiments demonstrate that AnchorRAG consistently outperforms existing baselines on four public QA datasets, establishing state-of-the-art results on the real-world reasoning tasks.
\end{itemize}

\section{Related Work}
Knowledge graphs\cite{bollacker2008freebase, auer2007dbpedia} provide essential supplementation to the static knowledge of LLMs\cite{gao2023retrieval}, effectively alleviating factual inaccuracies and hallucination issues during complex reasoning\cite{yih2016value} tasks. Consequently, the efficient retrieval of factual information from KGs for LLMs has become a critical topic of research, which can be divided into three main groups: \emph{Semantic Parsing}, \emph{Single-Pass Retrieval} and \emph{Iterative Retrieval} methods.

\subsection{Semantic Parsing Methods}
Semantic parsing methods aim to convert natural language questions into executable logical forms\cite{lan2022complex}, such as S-expressions or SPARQL queries. The results from these queries are then used as external knowledge to enhance the LLM's reasoning capabilities. Researchers have explored various strategies for generating and evaluating these logical forms. FlexKBQA\cite{li2024flexkbqa} first samples a multitude of S-expressions and employs a LLM to convert them into natural language questions, creating program-question pairs which are then used for training lightweight models. Another typical method RNG-KBQA\cite{ye-etal-2022-rng} resorts to a ranking model to select the most relevant logical forms based on a set of candidates. Additionally, ChatKBQA\cite{luo2023chatkbqa} introduces a ``Generate-then-Retrieve" framework, which first uses a fine-tuned LLM to generate a logical form skeleton and then populates it with specific details via an unsupervised retrieval module. It is important to note that if the generated logical forms are inaccurate, the entire reasoning pipeline will be confused, leading to incorrect final answers. This may limit the reasoning ability of semantic parsing methods in real-world scenarios, where accurate logical forms are hardly to generate. 

\subsection{Single-Pass Retrieval Methods}
Single-Pass retrieval methods enable the acquisition of all pertinent information, including entities, relations, or subgraphs, from a knowledge graph through a single retrieval operation\cite{peng2024GraphRAGsurvey, han2025GraphRAG}. These methods can effectively reduce the computational costs and reasoning latency. HippoRAG \cite{gutiérrez2025hippo} attempts to reduce the complexity and resources of retrieval process by constructing a schema-less KG, executing single-step retrieval guided by a personalized PageRank algorithm. Similarly, G-Retriever\cite{he2024g} formulates subgraph retrieval as a PCST optimization problem, creating a synergistic effect between the efficient retrieval and semantic prompting. Along this line, KnowGPT\cite{zhang2024knowgpt} leverages reinforcement learning to extract entire reasoning chains in a single pass. Furthermore, KG-GPT\cite{kim2023kggpt} generates a complete evidence graph in a single instance through sentence segmentation and graph retrieval. To sum up, through above methods simplify reasoning workflows and resources, they usually face inherent limitations in tackling complex multi-hop reasoning, where an initial retrieval that fails to identify critical evidence or filter the meaningless information may easily lead to generate incorrect answers.

\subsection{Iterative Retrieval Methods}
Iterative retrieval frameworks are designed to tackle complex queries by dynamically constructing and refining reasoning paths through multi-round ``retrieval-generation" operation\cite{zhang2025survey, han2025GraphRAG}. Early pioneering work, including ToG\cite{sun2023think} and StructGPT's IRR framework\cite{jiang2023structgpt}, demonstrates the viability of LLMs to progressively navigate KGs and accumulate evidence. To further handle the issues caused in the highly complex reasoning, recent research resorts to more sophisticated strategies, such as multi-stage optimization in GRAG\cite{hu-etal-2025-grag}, dynamic knowledge completion in Generate-on-Graph\cite{xu2024generate}, multi-agent deliberation in Generate-on-Graph\cite{ma2025debate}, and adaptive planning and backtracking mechanisms in Plan-on-Graph\cite{chen2024plan}. Despite their success, these advanced methods are fundamentally built on an idealized assumption: that entities within a query can be easily linked to their corresponding nodes in the knowledge graph. To extend to the open-world settings, some methods explore to apply entity linking strategy \cite{6823700, li-etal-2020-efficient, vollmers-etal-2025-contextual} to identify these entities. For instance, ToG\cite{sun2023think} and Paths-over-Graph\cite{tan2025paths} prompt LLM to automatically extract relevant entities, while AMAR\cite{xu2025amar} uses the ELQ\cite{li-etal-2020-efficient} model, which leverages a dual-encoder architecture to simultaneously detect mentions and link them to corresponding entities. However, these methods usually suffer from the issue caused by imprecise or partial matching, especially when the queries involving abbreviations, aliases, or long-tail entities. 


\section{Preliminary}
\textbf{Knowledge Graph} is commonly defined as a directed relational graph $\mathcal{G} = (\mathcal{E}, \mathcal{R}, \mathcal{T})$, where $\mathcal{E}$ denotes the set of entities, $\mathcal{R}$ denotes the set of relations, and $\mathcal{T} \subseteq \mathcal{E} \times \mathcal{R} \times \mathcal{E}$ is the set of factual triples. Each triple $(h, r, t) \in \mathcal{T}$ represents a factual statement, where $h \in \mathcal{E}$ is the head entity, $r \in \mathcal{R}$ is the relation, and $t \in \mathcal{E}$ is the tail entity. For example, a triple $(\textit{Barack Obama}, \textit{born\_in}, \textit{Hawaii})$ encodes the fact “Barack Obama was born in Hawaii.”

\textbf{Reasoning Path} is defined as a sequence of linked triples in $\mathcal{G}$ that connect a source entity $e_s$ to a target entity $e_t$. Formally, a path of length $L$ can be written as:
\begin{equation}\mathcal{P}=
\begin{Bmatrix}
(e_0,r_1,e_1),(e_1,r_2,e_2),\ldots,(e_{l-1},r_l,e_l)
\end{Bmatrix},\end{equation}
where $e_0 = e_s$ and $e_l = e_t$. Reasoning paths capture multi-hop dependencies that are crucial for multi-hop question answering.

\textbf{Reasoning Subgraph} is a connected subgraph $\mathcal{G}_q \subseteq \mathcal{G}$ constructed during the answering process for a query $q$. Typically, $\mathcal{G}_q = (\mathcal{E}_q, \mathcal{R}_q, \mathcal{T}_q)$ is iteratively expanded by retrieving the $k$-hop neighborhoods of frontier entities and refined to retain only the query-relevant triples. Such subgraphs serve as the evidence basis for reasoning.

\textbf{Knowledge Graph Question Answering (KGQA)} is the task of mapping a natural language query $q$ into one or multiple reasoning paths $\mathcal{P}$ or a reasoning subgraph $\mathcal{G}_q$, ultimately leading to the correct answer entity (or entities). The task requires both semantic understanding of the query and logical navigation in the KG.

\textbf{Open-World KG-RAG} extends the KGQA by relaxing the assumption that a set of ground-truth anchor entities $\mathcal{A} \subset \mathcal{E}$ is available for each query. Formally, given a natural language query $q$ containing a set of keywords $\mathcal{K} = \{k_1, k_2, \dots, k_n\}$, the Open-World KG-RAG is capable of autonomously learning a mapping function $\phi: \mathcal{K} \times \mathcal{G} \to \hat{\mathcal{A}}$, where $\hat{\mathcal{A}} \subset \mathcal{E}$ is the inferred candidate anchor set. Unlike the entity linking where the target entity is unique, Open-World KG-RAG is proposed for QA tasks where the target entity may not exist or may have multiple candidates.


\begin{figure*}[htbp]
\centering
\includegraphics[width=\linewidth]{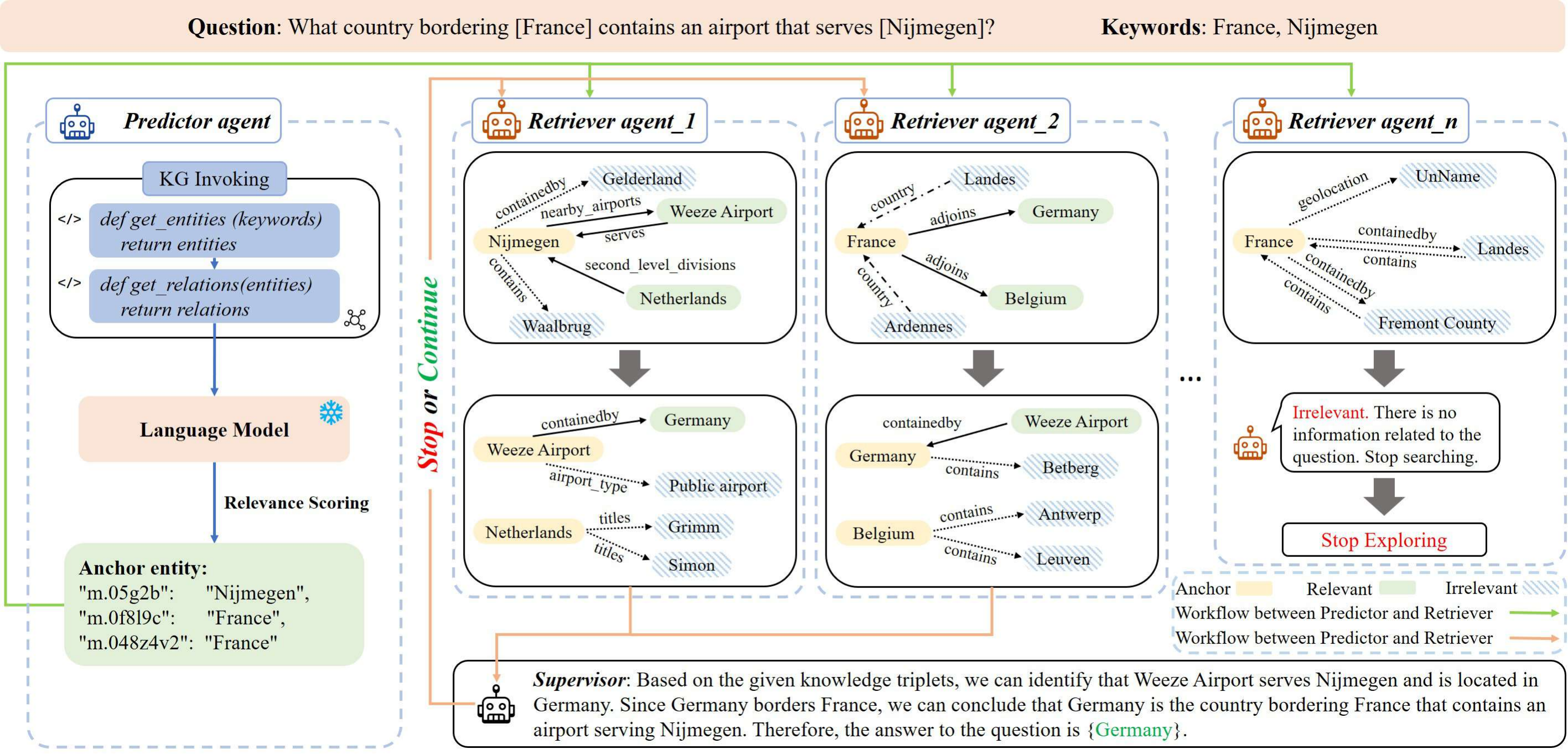}
\caption{An overview of the AnchorRAG framework, including the collaborative workflows of multi-agent exploration for question answering.}
\label{fig:framework}
\end{figure*}

\section{Method}

In this section, we propose AnchorRAG, an iterative retrieval RAG pipeline built upon a multi-agent collaborative framework, as shown in the Figure \ref{fig:framework}. The architectural design is primarily guided by the paradigms of ``divide and conquer" and ``collaborative refinement". We avoid the conventional single-threaded and error-prone ``link-then-retrieve" linear paradigm. Instead, we realize the complex open-world reasoning task with three specialized sub-tasks, each operated by a distinct agent. First, the Predictor Agent is responsible for identifying anchor entities within the user queries and aligning them with the knowledge graph. Then, the Retriever Agents perform multi-hop knowledge graph retrieval and filtering, parallel to cover critical information from multiple perspectives. Finally, the Supervisor Agent supervises the above retrieval processes and generates the answers based on the information retrieved from the given KG.. The prompt and in-context examples used in our method can be found in the Appendix \ref{app:prompts}.

\subsection{Anchor Entity Identifying}
In our framework, the process of accurately identifying anchor entities from natural language questions is handled by the predictor agent. This process consists of the following two steps.

\myparagraph{Candidate Entity Recognition}
The predictor agent is associated with a LLM to recognize and extract keywords from the input question $q$, forming a keyword set $\mathcal{W}=\{w_1,w_2,\ldots,w_n\}$, where each $w_i$ denotes an entity mention relevant to $q$. As shown in Figure \ref{fig:framework}, given the question $q$: ``What country bordering France contains an airport that serves Nijmegen?", the predictor extracts $\mathcal{W} = \{\text{France}, \text{Nijmegen}\}$. Considering the issues of entity abbreviations, aliases, long-tail entities, and potential spelling errors in open-world scenarios, directly using $\mathcal{W}$ for exact matching within a large-scale KG\cite{bollacker2008freebase} often results in low recall and high computational cost. To handle this issue, AnchorRAG employs a robust entity linking mechanism based on exact semantic indexing. Specifically, AnchorRAG encodes all entity names in the KG with MiniLM\cite{wang2020minilm}, then conducts a FAISS\cite{douze2024faiss} index from these embeddings, enabling efficient similarity search.
On this basis, Predictor first cleans and proofreads the initial set $\mathcal{W}$ of query keywords, and filters out irrelevant interferences such as non-critical terms, and is prompted to correct potential spelling errors. Subsequently, with the inverted file and product quantization techniques provided by the FAISS library, Predictor can efficiently retrieves a set of the most relevant candidate entities $\tilde{\mathcal{E}}$ from the knowledge graph.

\myparagraph{Context-Aware Entity Disambiguation}
The initial candidate set $\tilde{\mathcal{E}}$ often contains numerous entities with similar names that are irrelevant to the question's context. Taking all of them into account will increase the overhead of subsequent retrieval and inevitably introduce noise. AnchorRAG is inspired by the assumption that we can effectively determine the entity's relevance by measuring the semantic implied in its local graph structure. An entity is a more probable anchor if its neighboring relations are semantically aligned with the given question. Therefore, the Predictor employs a context-aware entity disambiguation strategy to precisely identify the most relevant candidates. Specifically, for each candidate entity $e \in \tilde{\mathcal{E}}$, the agent gathers its one-hop neighboring relations and obtains the set $\mathcal{R}(e)$. Then it invokes a pre-trained model SBERT \cite{reimers2019sentence} to obtain vector representations for the question $q$ and each relation $r_i \in \mathcal{R}(e)$, denoted as $\mathbf{q}$ and $\mathbf{r}_i$ respectively. The relevance score $s(r_i, q)$ between a relation and the question is then calculated using cosine similarity:

\begin{equation}
s(r_i, q) = \frac{\mathbf{r}_i \cdot \mathbf{q}}{|\mathbf{r}_i| |\mathbf{q}|},
\end{equation}

Furthermore, we define the final relevance score for the candidate entity $e$ by aggregating the scores of its most relevant neighboring relations:

\begin{equation}
\mathrm{Score}(e, q) = \frac{1}{k} \sum_{r \in \mathcal{N}(e)} s(r, q),
\end{equation}
where $\mathcal{N}(e)$ denotes the set of top-$k$ relevant relations according to the Eq.(1). By scoring and ranking all candidates in the set $\tilde{\mathcal{E}}$, the Predictor selects the top-$m$ entities with the higher relevance scores (e.g., France and Nijmegen) as the final anchor set, which will be passed to the following retriever agents for parallel searching.

\subsection{Parallel Exploration}
With the top-$m$ candidate anchor entities $\{a_1, a_2, \dots, a_m\}$ obtained by the Predictor agent, the framework initiates a parallel exploration operation, assigning each candidate to an independent retriever agent $\mathsf{R}_i$. This multi-agent parallel search is fundamental to the robustness of our framework, as it alleviates the issue of linking to a single, potentially erroneous anchor. Instead of following a linear path, AnchorRAG explores multiple reasoning hypotheses simultaneously. Each retriever agent $\mathsf{R}_i$ maintains a reasoning subgraph $G_i^t = (V_i^t, E_i^t)$ at iteration $t$, initialized as $G_i^{(0)} = ({a_i}, \emptyset)$.
At each iteration, the agent expands its current exploration frontier by retrieving the one-hop neighborhoods of its active nodes, forming a candidate subgraph. To maintain relevance and reduce overhead, this candidate subgraph then undergoes a two-stage refinement process. First, a candidate ranking step scores and selects the most promising relations and entities for expansion. On this basis, an evidence validation step assesses the generated triples, retaining only those that include relevant evidence for the query. The entities within the validated subgraph are considered as the new frontier nodes for the next iteration of expansion.

\myparagraph{Candidate Relations Ranking}
At the $t$-th retrieval iteration, each retriever agent dynamically identifies the most relevant relational paths from the active entity $e_i^t$ through semantic alignment. Given the knowledge graph $\mathcal{G}$, the agent first retrieves all one-hop relations connected to $e_i^t$, denoted as $\mathcal{R}{(e_i^t)} = \{r_1, r_2, \dots, r_k\}$.Then a LLM acts as a semantic filter that assigns a relevance score $s_{\text{r}}(r_j \mid q, e_i^t)$ to each candidate relation $r_j \in \mathcal{R}_{e_i^t}$. The prompt used for this scoring integrates the query context, the current entity, and all retrieved relations, allowing the model to perform context-aware semantic alignment. Subsequently, the Retriever selects the top-$b$ relations with the highest scores, forming a refined relation set $\tilde{\mathcal{R}}_{e_i^t}$. As illustrated in Figure~\ref{fig:framework}, for the given query, Retriever $\mathsf{R}_1$ starts from the anchor entity ``$\mathbf{Nijmegen}$" and identifies highly relevant relations such as ``$\mathbf{nearby\_airports}$", ``$\mathbf{serves}$", and ``$\mathbf{second\_level\_divisions}$". Retriever $\mathsf{R}_2$ and Retriever $\mathsf{R}_3$ start from "$\mathbf{France}$", and then judge that relations like ``$\mathbf{adjoins}$" and ``$\mathbf{containedby}$" are beneficial to exploring the answer, while filtering out irrelevant relations such as ``$\mathbf{contains}$". During this process, AnchorRAG demonstrates strong generality across diverse knowledge graphs without requiring any training or fine-tuning of models.


\myparagraph{Candidate Entities Ranking}
At the $t$-th iteration, for each selected relation $r_j \in \tilde{\mathcal{R}}{e_i^t}$, the retriever agent $\mathsf{R}_i$ traverses $\mathcal{G}$ to obtain all connected entities, forming a candidate entity set $\mathcal{E}({r_j}) = \{ e_1, e_2, \dots, e_{n_j} \}$. Note that directly evaluating individual entities without considering their relational context may be suboptimal: such approach may inevitably filter out intermediate entities that are crucial for multi-hop reasoning. To address this problem, $\mathsf{R}_i$ employs a LLM-based scoring mechanism that evaluates the contextual relevance $s_{\text{e}}(e_k \mid q, e_i^t, r_j)$ for each entity $e_k \in \mathcal{E}_{r_j}$. The structured prompt integrates the query $q$, the current anchor entity $e_i^t$, and the relation $r_j$, allowing the model to perform fine-grained semantic reasoning. All candidate entities are then ranked according to their scores, and for each relation $r_j$, the top-$b$ entities are retained:

\begin{equation}\tilde{\mathcal{E}}_{r_j}=\mathrm{Top}_b(\mathcal{E}_{r_j},s_{\mathrm{e}})\end{equation}

These selected entities will form the preliminary reasoning paths $\mathcal{T}(e_i^t)$, which serve as the evidence for the subsequent validation phase.

In addition, to reduce computational cost, we propose an alternative hybrid retrieval strategy that replaces the LLM-based scoring with a combination of sparse (e.g., BM25) and dense (e.g., SBERT) retrieval models. This hybrid approach allows AnchorRAG to efficiently recall semantically or lexically relevant candidates, balancing efficiency and reasoning performance. The effectiveness of this variant is further evaluated in the efficiency evaluation section.


\myparagraph{Evidence Validation}
In the candidate ranking stage, each candidate relation $r_j$ and entity $e_k$ is assigned a relevance score $s_{\text{r}}(r_j)$ and $s_{\text{e}}(e_k)$, respectively. Existing methods usually apply a linear combination of these two relevance measures to approximate the similarity of the reasoning path formed by the relation–entity pair. They implicitly assume that the relevance of relations and entities is independent, while neglecting their inherent semantic coherence. As a result, they fail to capture the complex semantic interactions and implicit logical structures within the knowledge graph, and thus limit the retrieval performance. To further locate query-relevant facts hidden in the knowledge graph, AnchorRAG introduces a novel evidence validation mechanism. Specifically, each $\mathsf{R}_i$ converts its candidate reasoning paths $\mathcal{T}(e_i^t)$ into natural language text sequences, denoted as $\mathcal{X}_i^t = \{x_1, x_2, \dots, x_{p_i}\}$, including both the query $q$ and contextual information. On this basis, the agent leverages the reasoning and comprehension capabilities of a LLM to identify relevant paths that contain critical evidence, and filter out irrelevant or inconsistent evidence. As illustrated in Figure~\ref{fig:framework}, Retriever $\mathsf{R}_1$ validates the reasoning chain $\mathcal{P}_1$ = (\textbf{Nijmegen, nearby\_airports, Weeze Airport}), (\textbf{Weeze Airport, containedby, Germany}), while Retriever $\mathsf{R}_2$ identifies another path $\mathcal{P}_2$ = (\textbf{France, adjoins, Germany}), (\textbf{Weeze Airport, containedby, Germany}). These validated subgraph $\tilde{G}_i^t$ is then stored in a shared global memory $\mathcal{M}$ accessible to all agents, serving as an important source for subsequent reasoning iterations and answer generation.
 

\begin{table*}[tb]
\centering
\small
\renewcommand{\arraystretch}{1.4}
\setlength{\tabcolsep}{14pt} 
\setlength{\aboverulesep}{0pt}
\setlength{\belowrulesep}{0pt}
\begin{tabular}{lcccccccc}
\toprule[1.pt]
\multirow{2}{*}{\textbf{Dataset}} & \multicolumn{2}{c}{\textbf{WebQSP}} & \multicolumn{2}{c}{\textbf{GrailQA}} & \multicolumn{2}{c}{\textbf{CWQ}} & \multicolumn{2}{c}{\textbf{WebQuestions}} \\
\cmidrule(lr){2-3}
\cmidrule(lr){4-5}
\cmidrule(lr){6-7}
\cmidrule(lr){8-9}
& Hit@1 & Acc & Hit@1 & Acc & Hit@1 & Acc & Hit@1 & Acc \\ \midrule
Method & \multicolumn{8}{c}{\textbf{Qwen-Plus}} \\ \midrule
\textbf{IO} & 63.3 & 42.2 & 33.2 & 26.6 & 33.2 & 33.2 & \underline{56.3} & 43.2 \\
\textbf{Chain-of-Thought} & 62.9 & 41.5 & 32.3 & 26.7 & 37.6 & 37.6 & 54.1 & 41.2 \\
\textbf{Self-Consistency} & 63.0 & 41.2 & 33.8 & 28.4 & 39.2 & 39.2 & 52.7 & 40.1 \\

\textbf{PoG} & 33.1 & 25.4 & 36.9 & 33.0  & 39.5 & 29.2 & 25.3 & 21.5 \\
\textbf{ToG} & \underline{66.1} & \underline{46.3} & \underline{41.9} & \underline{35.3} & \underline{39.8} & \underline{39.8} & 56.1 & \underline{43.7} \\
\textbf{AnchorRAG(Ours)} & \textbf{73.3} & \textbf{56.4} & \textbf{62.7} & \textbf{56.0} & \textbf{47.0} & \textbf{47.0} & \textbf{60.9} & \textbf{50.5} \\ \midrule
Method & \multicolumn{8}{c}{\textbf{GPT-4o-mini}} \\ \midrule
\textbf{IO} & 65.8 & 46.1 & 35.6 & 27.9 & 35.7 & 35.7 & 57.7 & 45.4 \\
\textbf{Chain-of-Thought} & 62.5 & 41.6 & 31.0 & 25.8 & 34.5 & 34.5 & 53.6 & 41.2 \\
\textbf{Self-Consistency} & 59.2 & 40.1 & 34.0 & 27.1 & 36.1 & 36.1 & 50.7 & 39.3 \\
\textbf{PoG} & 55.3 & 36.4 & 39.0  & 34.2 & 36.1 & 36.1 & 41.8  & 33.3 \\
\textbf{ToG} & \underline{71.4} & \underline{49.9} & \underline{48.7} & \underline{40.9} & \underline{40.7} & \underline{40.7} & \textbf{61.4}  & \underline{47.6} \\
\textbf{AnchorRAG(Ours)} & \textbf{74.1} & \textbf{57.4} & \textbf{63.4} & \textbf{56.8} & \textbf{44.8} & \textbf{44.8} & \underline{60.0} & \textbf{50.0} \\
\specialrule{1pt}{0pt}{0pt}
\end{tabular}
\caption{The results of different methods on various Datasets, using Qwen-Plus and GPT-4o-mini as the LLM. The best results are highlighted in bold. The suboptimal results are underlined.}
\label{tab:overall}
\end{table*}

\subsection{Reasoning}
After the parallel exploration phase, each retriever agent $\mathsf{R}_i$ has obtained a reasoning subgraph $G_i^t$ (and its validated subset $\tilde{G}_i^t$) which are stored in the shared global memory $\mathcal{M}$. The Supervisor agent $\mathcal{S}$ aggregates these multi-perspective evidences and checks whether the obtained information is sufficient to answer the query $q$. If the evidence chain is complete and results in a clear conclusion, the Supervisor generates the answer directly. If the evidence is insufficient, Supervisor $\mathcal{S}$ evaluates all candidate reasoning paths and drives the high relevance Retriever to explore in a promising direction at the next round, which are marked as $[\mathbf{active}]$. Conversely, retrievers deemed to deviate from the query or retrieve noisy, irrelevant information are terminated early, reducing the overhead and limiting the influence of irrelevant data in the global memory.
Exploration stops under either of the following two conditions: (1) a predefined maximum search depth $\mathcal{D}$ is reached, or (2) no retrievers remain in the $[\mathbf{active}]$ state. In such cases, the Supervisor concludes that further KG-based path exploration is unlikely to generate an answer and applies the chain-of-thought (CoT)\cite{wei2022chain} reasoning mechanism instead to directly obtain the answers.


\section{Experiments}
\myparagraph{Dataset and Evaluation}
In the experiments, we adopt four benchmark datasets to evaluate the performance of our framework, including three multi-hop Knowledge Graph Question Answering (QA) datasets: WebQuestionSP (WebQSP) \cite{yih2016value}, GrailQA \cite{gu2021beyond} and Complex WebQuestions (CWQ) \cite{talmor2018web}, and one for Open-Domain Question Answering: WebQuestions \cite{berant2013semantic}. Freebase \cite{bollacker2008freebase} is utilized as the external Knowledge Graph, which contains the complete knowledge facts that are necessary to answer questions in the above datasets. We select Freebase to ensure fair comparison with established benchmarks\cite{sun2023think, chen2024plan} whose structural complexity and large scale poses a challenge to evaluating the reasoning capabilities of RAG methods. In addition, we apply the exact match (i.e., Hit@1) and accuracy (i.e., Acc) as evaluation metrics for our method which are widely used in the field of question answering with RAG.

\myparagraph{Implementation Settings}
We extract all entities from Freebase, encode their name description into vectors using SBERT model (all-MiniLM-L6-v2, without fine-tuning) \cite{reimers2019sentence}, and subsequently index them in a vector database \cite{douze2024faiss} for downstream entity querying and matching. The knowledge graph derived from Freebase is deployed on a locally deployed Virtuoso \cite{erling2009rdf} server for our experiments. We selected GPT-4o-mini and Qwen-Plus as the fundamental LLM in our framework. To prevent redundant exploration by agents and reduce computational overhead, we set both the maximum search depth $\mathcal{D}$ and the expansion width $b$ to 3. This configuration aligns with the empirically optimal values identified in \cite{sun2023think}. Additionally, we fix the number of neighbor relations $k$ during anchor entity identifying to 5, and the number of retrieval agents $m$ is set to 3 for parallel multi-hop reasoning.

\myparagraph{Baselines}
Some widely used prompting strategies are selected as the baselines, including vanilla IO prompting \cite{brown2020language}, chain-of-Thought (CoT) \cite{wei2022chain}, and Self-Consistency \cite{wang2022self}. Moreover, we also compare the proposal with two representative RAG methods: Think-on-graph \cite{sun2023think} and Plan-on-graph \cite{chen2024plan}. It is worth noting that the above two methods are both implemented in the closed-world setting where the anchor entities have been pre-defined and are part of the annotated datasets. Thus, to test their performance in the specific scenario where the anchor entity is directly available, we apply the fundamental LLM to select the anchor entities for the following open-world question answering. Other settings and parameters remain unchanged.

\begin{table}[t]
\centering
\renewcommand{\arraystretch}{1.0} 
\setlength{\tabcolsep}{4.5pt}
\begin{tabular}{lcccc}
\toprule
\multirow{2}{*}{\textbf{Dataset}} & \multicolumn{2}{c}{\textbf{WebQSP}} &  \multicolumn{2}{c}{\textbf{GrailQA }} \\
\cmidrule(lr){2-3}
\cmidrule(lr){4-5}
& normal & open-world & normal & open-world  \\
\midrule
Method & \multicolumn{4}{c}{\textbf{Qwen-Plus}} \\ \midrule
ToG   & 66.1 & 58.5$\downarrow$\ts{11.5\%} & 41.9 & 32.1$\downarrow$\ts{23.4\%} \\
AnchorRAG  & 73.3 & 71.3$\downarrow$\ts{2.7\%} & 62.7 & 55.4$\downarrow$\ts{11.6\%} \\
\midrule
Method & \multicolumn{4}{c}{\textbf{GPT-4o-mini}} \\ \midrule
ToG   & 71.4 & 67.1$\downarrow$\ts{6.0\%} & 43.7 & 35.5$\downarrow$\ts{18.8\%} \\
AnchorRAG  & 74.1 & 70.3$\downarrow$\ts{5.1\%} & 63.4 & 55.4$\downarrow$\ts{12.6\%} \\
\bottomrule
\end{tabular}
\caption{Performance comparisons (Hits@1) of different methods on the normal and open-world versions of WebQSP and GrailQA datasets, using Qwen-Plus and GPT-4o-mini as the backbone LLMs.}
\label{tab:open-world qa}
\end{table}

\subsection{Main Results}
The main experimental results, as shown in Table \ref{tab:overall}, indicate that our proposed method achieves a significant performance advantage over most baseline methods when deployed with both the Qwen-Plus and GPT-4o-mini LLMs. Specifically, when applying Qwen-Plus, AnchorRAG achieves an average improvement by 10\% and 11.2\% on the metrics
Hit@1 and Accuracy respectively, compared to the strongest baseline. Besides, with GPT-4o-mini model, our proposal also achieves the increases of 6.3\% and 8.6\%. We can also find that our method obtain an outstanding performance on the complex datasets like GrailQA, where its Hit@1 and Accuracy scores are increased substantially by 20.8\% and 20.7\%, respectively. The reason may be that the questions and the KG entities are semantically inconsistent in this dataset, and the anchor entities are more difficult to locate in this case. AnchorRAG can handle this issue by the multi-agent collaboration, thereby obtaining the state-of-the-art results. In addition, the performance of the RAG methods, e.g., ToG, consistently surpasses the methods solely relying on the internal knowledge, e.g., CoT. This demonstrates the effectiveness of the RAG methods in the question answering task, especially when dealing with the complex reasoning such as GrailQA and CWQ. 

\subsection{Robust Analyses}
Since existing public QA datasets primarily assume that the external KG contains all the necessary information and rely on clean entity linking, they are ill-suited for the open-world question answering task. Thus, to further evaluate the generality and robustness of our method, we constructed the open-world versions of the WebQSP and GrailQA datasets. Specifically, we first added semantic noise by introducing typos into the original questions. These perturbations can effectively simulate real-world scenarios, thereby testing the capability of the proposed method for the open-world RAG. Then, we incorporated additional questions into the datasets whose answers could not be directly retrieved from the KG. The detailed construction process is provided in the appendix \ref{app:Details of Experimental Datasets}. The results on the self-constructed datasets are presented in Table \ref{tab:open-world qa}, where we can observe that our method shows strong robustness on this open-world scenario. For instance, on GrailQA, the performance of ToG using Qwen-Plus drops by 23.4\%, whereas our method only drops by 11.6\%. These results demonstrate that AnchorRAG can effectively address the open-world QA tasks and further enhance the generality of the RAG methods.

\begin{table}[t]
\centering
\renewcommand{\arraystretch}{1.0} 
\setlength{\tabcolsep}{12pt}
\begin{tabular}{lcc}
\toprule
\textbf{Method}        &    WebQSP  &   GrailQA  \\
\midrule
\textbf{AnchorRAG}     &   74.1    &   63.4      \\
w/o Entity Disambiguation     &    67.8     &    42.8     \\
w/o Parallel Exploration     &   72.8    &     60.2    \\
w/o LLM Ranking             &  71.1    &  64.2  \\
w/o Evidence Validation    &  71.6   &   62.8  \\
\bottomrule
\end{tabular}
\caption{Ablation study of AnchorRAG: Effect of different components on the Hits@1 performance.}
\label{tab:ablation}
\end{table}

\subsection{Ablation Studies}
To validate the effectiveness of the core components within our proposed AnchorRAG framework, we conducted the following ablation studies on the WebQSP and GrailQA datasets, with the results presented in Table \ref{tab:ablation}. Note that ``w/o Entity Disambiguation" removes the neighborhood information during anchor entity identification, ``w/o Parallel Exploration" eliminates the multi-agent retrieval mechanism, replacing it with a single-path retrieval approach. The ``w/o LLM Ranking" variant (referred to as \textbf{AnchorRAG-LR} in the efficiency evaluation subsection) uses a lightweight hybrid retrieval strategy instead of LLM ranking. It works by combining SBERT semantic similarity and BM25 scores for efficient candidate filtering. And ``w/o Evidence Validation" refers to the absence of fine semantic filtering for candidate triples during retrieval. The results show that each component can contribute to the question answering performance. Comparing with parallel exploration and evidence validation, the entity disambiguation module has a greater impact on the performance, achieving the improvement by 6.3\% and 20.6\% on WebQSP and GrailQA respectively, which emphasizes the importance of precise anchor entities in the open-world question answering tasks.
Interestingly, the ``w/o LLM Ranking" variant performs slightly better than the AnchorRAG on GrailQA. This is likely because GrailQA contains more structured and well-aligned relations, where lightweight semantic retrieval (SBERT + BM25) can efficiently the semantic associations without the excessive reasoning conducted by the LLM.

\subsection{Efficiency evaluation}
\begin{figure}[tbp]
    \makebox[\linewidth][c]{%
        \includegraphics[width=1\linewidth]{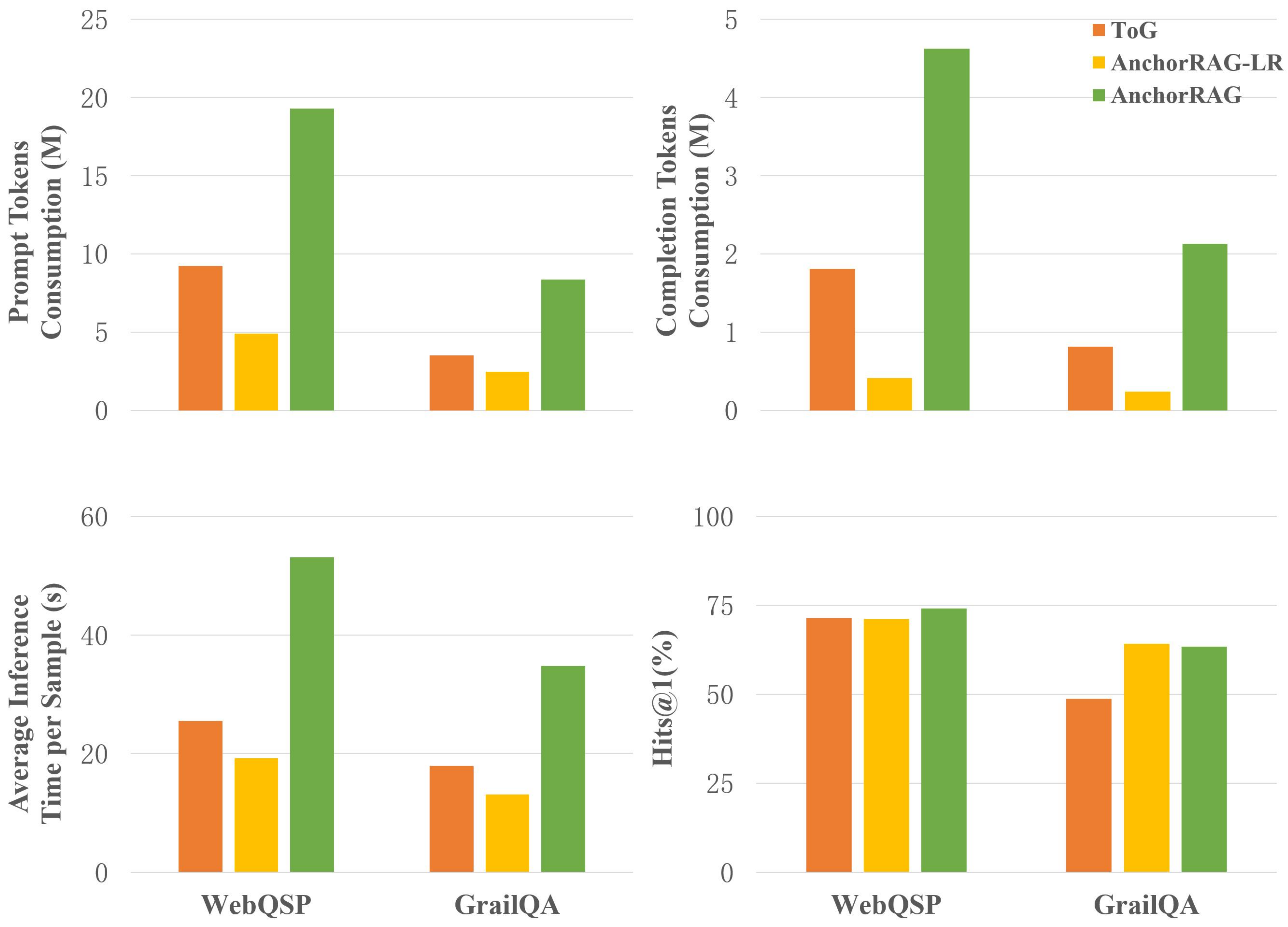}%
    }
	\caption{Efficiency and Performance Comparison of Different Methods on the WebQSP and GrailQA Datasets, using GPT-4o-mini as the backbone LLM.}
	\label{fig:Efficiency evaluation}
\end{figure}

In this efficiency evaluation, we compared the performance of AnchorRAG, AnchorRAG-LR along with the strongest baseline ToG in terms of computational efficiency and answer quality. The results are shown in Figure \ref{fig:Efficiency evaluation}, where ``Prompt Tokens Consumption (M)" represents the overhead of input tokens that formulate the LLM prompt, ``Completion Tokens Consumption (M)" indicates the number of tokens generated in the LLM responses and ``Average Inference Time per Sample (s)" measures the average reasoning time for answering each query. From the results, we can obviously observe that AnchorRAG-LR achieves substantially faster reasoning than both ToG and AnchorRAG, while also effectively reducing the overhead of the prompt or completion tokens. For example, on the WebQSP dataset, AnchorRAG-LR reduces the total token consumption by 48.2\% compared to ToG and by 77.8\% compared to AnchorRAG; its average reasoning time is only 75.3\% and 36.2\% of that of the methods ToG and AnchorRAG. Similarly,  we can observe that on the GrailQA dataset, where token consumption decreases by around 37.4\% and 74.2\%, and reasoning speed improves by 26.8\% and 62.4\%, respectively. Additionally, in terms of answer accuracy metrics, AnchorRAG remains superior due to its LLM-based powerful reasoning capability. In summary, AnchorRAG achieves stronger semantic performance at a higher computational cost, whereas AnchorRAG-LR provides a practical and efficient alternative in the resource-constrained scenarios. In addition, we recommend prioritizing the LLM-ranking strategy (AnchorRAG) for those highly ambiguous queries that cannot be easily tackled by lightweight models, e.g., SBERT.

\subsection{Effect of the Retrieval}
\begin{figure}[t]
    \makebox[\linewidth][c]{%
        \includegraphics[width=1\linewidth]{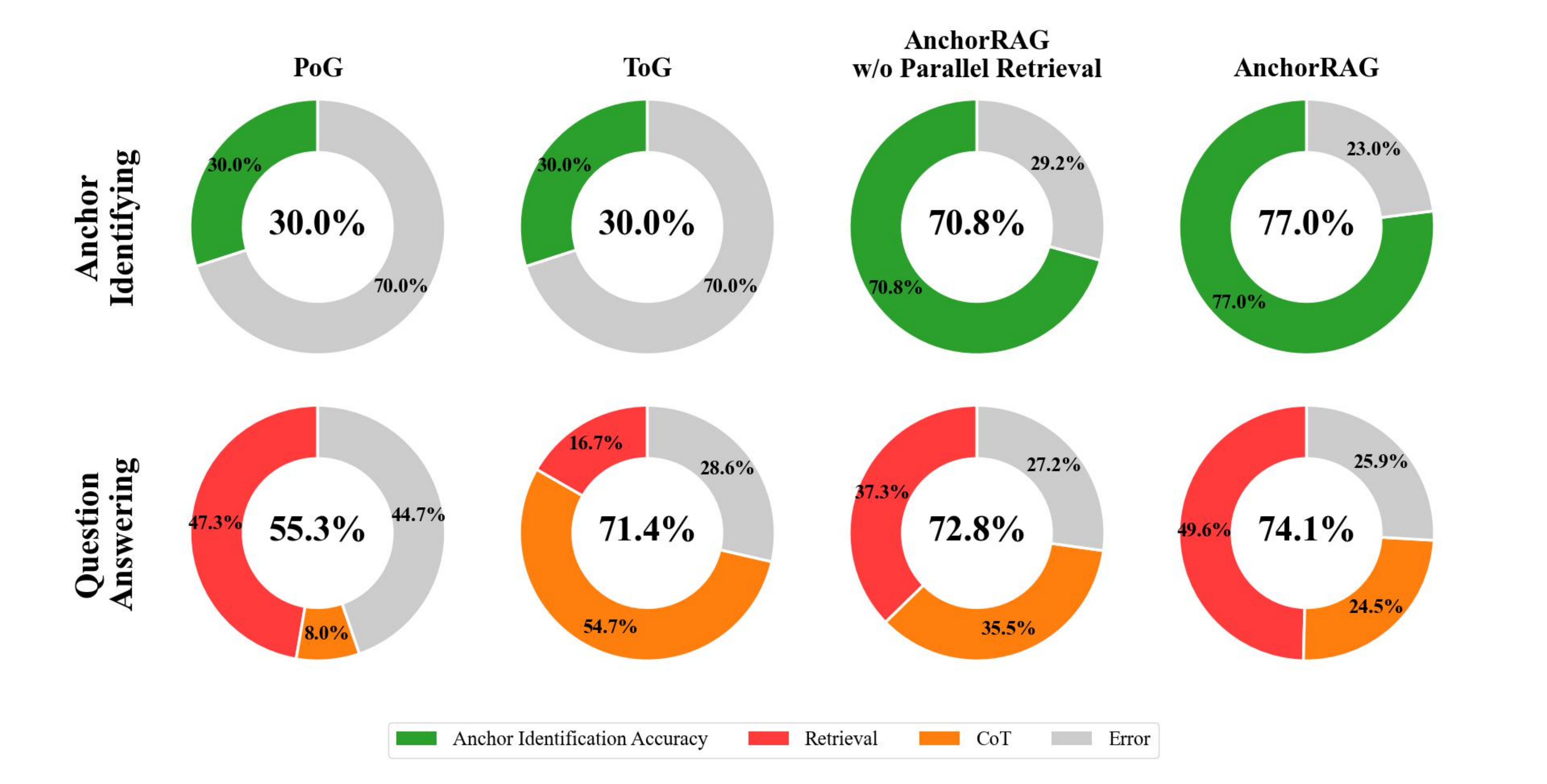}%
    }
	\caption{The effect of the anchor identifying and knowledge retrieval on WebQSP.}
	\label{fig:Retrieval Performance}
\end{figure}

To further validate the effectiveness of our method in anchor identifying and knowledge retrieval, we compare   the effect of retrieval of our method against two RAG baselines (PoG and ToG) and a variant (AnchorRAG w/o Parallel Retrieval). The results are shown in Figure \ref{fig:Retrieval Performance}. The figures in the top row illustrate the accuracy of anchor identifying (marked in green), where our method significantly outperforms the baselines by disambiguating the entities with their textual and neighborhood information. The bottom row presents the proportion of the exact match for the answers, including the answers obtained by retrieval (marked in red) and those inferred by CoT reasoning (marked in orange). Notably, AnchorRAG achieves the highest retrieval rate at 49.6\%, indicating that the multi-agent collaborative mechanism can effectively obtain the correct answer entities, which supports the following reasoning process to improve the QA performance.

\subsection{Hyper-parameter Analyses}
We conducted the parameter sensitivity analyses on GrailQA to investigate the optimal configuration for the number of neighborhood relations $k$ and retrieval agents $m$, where the results are shown in Table \ref{tab:Hyper-parameter Analysis}. We can find that retaining excessive neighboring relations $k$ for each candidate entity will lead to a performance loss, due to the fact that it may introduce noise and neglect the key structural information. 
\begin{table}[t]
\centering
\renewcommand{\arraystretch}{1.0} 
\setlength{\tabcolsep}{12pt}
\begin{tabular}{cccc}
\toprule
\textbf{Setting} & $k=3$ & $k=5$ & $k=7$ \\
\midrule
$m=1$ &   60.4  &   57.6   &   56.4  \\
$m=2$ &   63.6  &   63.3   &   61.0  \\
$m=3$ &   63.1  &   63.4   &   63.3  \\
$m=4$ &   62.6  &   63.1   &   62.4  \\
\bottomrule
\end{tabular}
\caption{The results of AnchorRAG with different hyper-parameters using GPT-4o-mini on GrailQA.\vspace{-1em}}
\label{tab:Hyper-parameter Analysis}
\end{table}
Besides, a proper number of retriever agent $m$ can make AnchorRAG achieve the optimal results, which can reflect the effectiveness of the MAC framework. In addition, we can also find that more retrieval agents won't always obtain the better results. Excessive agents will introduce uncertain retrieval information into the collaborative process, thereby impairing the performance. Furthermore, we also conduct the hyperparameter sensitivity analysis for retrieval depth and width. The detailed experimental results and analyses can be found in the appendix  \ref{app:additional Hyper-parameter}.

\section{Conclusion}
In this paper, we present AnchorRAG, a novel multi-agent collaborative RAG framework for open-world Knowledge Graph Question Answering. Unlike existing methods that rely on predefined anchor entities, AnchorRAG overcomes the bottleneck of unreliable entity linking by identifying and grounding plausible anchors without prior assumptions.  Specifically, a predictor agent dynamically identifies candidate anchor entities by aligning the given questions with the candidate entities and initializes independent retriever agents to conduct parallel multi-hop explorations. Then a supervisor agent formulates iterative retrieval strategy for these retriever agents and synthesizes the reasoning paths to generate the final answer. Extensive experiments on four public KGQA benchmarks demonstrate that AnchorRAG significantly outperforms the state-of-the-art baselines, validating the effectiveness of our MAC framework, especially when dealing with the complex multi-hop reasoning tasks.

\begin{acks}
This work was supported  in
part by the National Natural Science Foundation of China (62476274, U22B2048, 62394330), in part by the Special Project of Joint Funding for Municipal, University (Institute) and Enterprise under the Guangzhou Basic Research Program (2024A03J0395),  in part by the Science and Technology
Projects in Guangzhou (2024D03J0010).
\end{acks}

\bibliographystyle{ACM-Reference-Format}
\balance
\bibliography{main}

\appendix

\section{Details of Experimental Datasets}
\label{app:Details of Experimental Datasets}
We use four benchmark datasets: WebQuestionSP (WebQSP) \cite{yih2016value}, GrailQA \cite{gu2021beyond} and Complex WebQuestions (CWQ) \cite{talmor2018web}, and WebQuestions (WebQ) \cite{berant2013semantic} to evaluate the effectiveness of our method. These datasets include a training set, a validation set, and a test set. For fair comparison, we follow the baselines \cite{sun2023think, chen2024plan} and use only the test sets as evaluation benchmarks. 
\begin{table}[bp]
\centering
\renewcommand{\arraystretch}{1.2} 
\setlength{\tabcolsep}{6pt}
\begin{tabular}{lccc}
\toprule
\textbf{Datasets} & \#Train & \#Validation & \#Test \\
\midrule
WebQSP &   3098  &   -   &   1639  \\
GrailQA &  44,337   &   6763   &   13231  \\
CWQ &   27734  &   3480   &   3475  \\
Webquestions &  3778   &   -   &   2032  \\
WebQSP(Open-World) & 3098 & - & 1803 \\
GrailQA(Open-World) & 44,337 & 6763 & 1100 \\
\bottomrule
\end{tabular}
\caption{Statistics of datasets.\vspace{-1em}}
\label{tab:datasets}
\end{table}
\begin{table}[bp]
\centering
\renewcommand{\arraystretch}{1.2} 
\setlength{\tabcolsep}{12pt}
\begin{tabular}{lccc}
\toprule
\textbf{Datasets} & 1 hop & 2 hop & $\ge$ 3 hop \\
\midrule
WebQSP &   65.49\%  &   34.51\%   &   0.00\%  \\
GrailQA &  68.92\%   &   25.82\%   &   5.26\%  \\
CWQ &   40.91\%  &   38.34\%   &   20.75\%  \\
\bottomrule
\end{tabular}
\caption{statistics of the reasoning hops in WebQSP, CWQ and GrailQA.\vspace{-1em}}
\label{tab:hop}
\end{table}
In order to save computational resources, following prior research \cite{sun2023think, chen2024plan}, we only used a sample of 1000 data points from the GrailQA dataset for evaluation. Noticing that, in order to verify the performance of each method in open world settings, we introduce noise into the WebQSP and GrailQA datasets, rewriting keywords in the questions to make it impossible to directly match entities in the knowledge graph through string matching. We also sample a small portion of questions from two other datasets, HotpotQA \cite{yang2018hotpotqa} and TriviaQA \cite{joshi2017triviaqa}, and merge them with the noisy data from WebQSP and GrailQA to create two open world datasets (Open-World WebQSP and Open-World GrailQA) to verify the robustness of our method. The statistics of the datasets are shown in Table \ref{tab:datasets}, and the statistics of the reasoning hops are shown in Table \ref{tab:hop}.

\section{Experiments of Additional Hyper-parameter}
\begin{figure}[h!]
    \makebox[\linewidth][c]{%
        \includegraphics[width=1.0\linewidth]{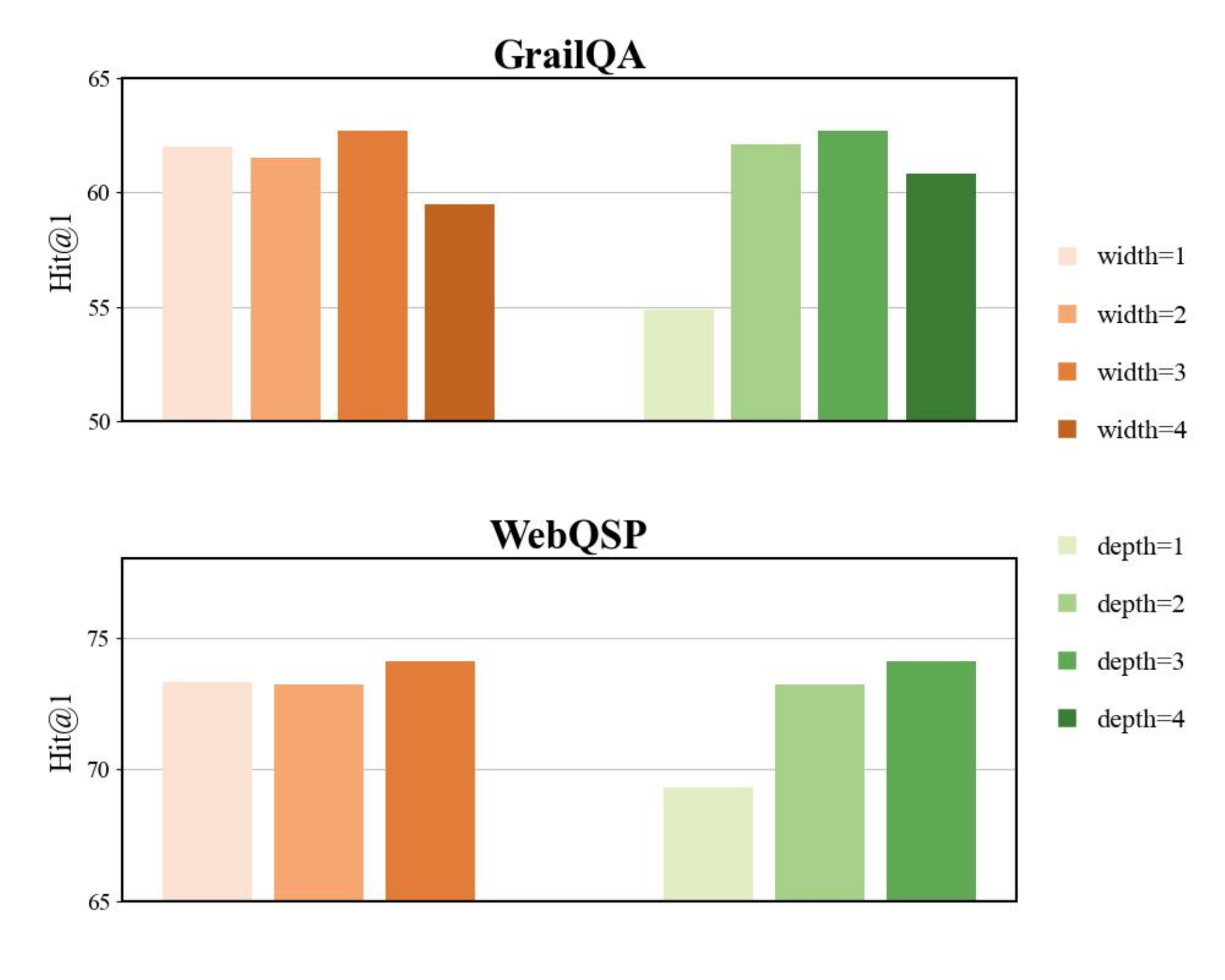}%
    }
	\caption{Performances of AnchorRAG with different search depths and widths.}
	\label{fig:width and depth}
\end{figure}

Figure \ref{fig:width and depth} shows the hyperparameter sensitivity analysis for retrieval depth and width. The experimental results strongly validate the effectiveness of our default configuration, i.e., width=3 and depth=3. On both datasets, the Hit@1 performance of our method peaks when these settings are used. This non-linear performance trend is attributed to two reasons. First, the performance improvement is limited by the inherent complexity of the datasets, as most questions do not require more than 3 hops of reasoning, as shown in table \ref{tab:hop}. Therefore, a deeper search does not yield significant benefits. Second, excessively expanding the search space introduces a significant amount of noise from irrelevant entities and relations, which have the negative impact on the reasoning and lead to a poorer performance. This is shown by the performance degradation on GrailQA when depth or width is increased to 4. In particular, the WebQSP dataset does not contain questions requiring more than 3 hops. Consequently, increasing the retrieval depth beyond 3 on WebQSP offers no performance benefit, and we therefore did not perform experiments with depth=4 on this dataset. To sum up, our parameter selection presents an optimal trade-off between maintaining adequate search space, mitigating noise, and balancing computational overhead.

\label{app:additional Hyper-parameter}

\section{Experiments on Additional LLM Backbones}
To further evaluate the universality and robustness of AnchorRAG with different Large Language Model (LLM) backbones, we extend our experiments using two additional state-of-the-art models: \textbf{Claude4.5-haiku} and \textbf{Gemini-2.5-flash}. We compared the performance of our method against the strongest baseline, i.e., ToG, on the GrailQA dataset. The experimental results are presented in Table \ref{tab:additional_models}. Consistent with the results in the main experiments, AnchorRAG demonstrates superior performance over the ToG on both new backbones. For example, on the GrailQA dataset, AnchorRAG achieves substantial improvements in Hit@1, surpassing ToG by \textbf{13.7\%} with Claude4.5-haiku and \textbf{16.5\%} with Gemini-2.5-flash. This significant improvement validates that our multi-agent collaboration framework effectively mitigates the entity linking issues in complex open-world scenarios, regardless of the underlying LLM. These findings suggest that AnchorRAG's effectiveness is model-agnostic and can cooperate with various LLMs to enhance open-world reasoning capabilities.

\begin{table}[t]
\centering
\renewcommand{\arraystretch}{1.3} 
\setlength{\tabcolsep}{15pt}
\begin{tabular}{lccc}
\toprule
\multirow{2}{*}{\textbf{Dataset}} &  \multicolumn{3}{c}{\textbf{GrailQA }} \\
\cmidrule(lr){2-4}
& Hit@1 & Acc & F1   \\
\midrule
\multicolumn{4}{c}{\textbf{Claude4.5-haiku}} \\ \midrule
ToG   & 31.7 & 26.6 & 27.6 \\
AnchorRAG  & \textbf{45.4} & \textbf{39.0} & \textbf{40.3} \\
\midrule
\multicolumn{4}{c}{\textbf{Gemini-2.5-flash}} \\ \midrule
ToG   & 39.6 & 32.8 & 34.0 \\
AnchorRAG  &\textbf{56.1} & \textbf{47.9} & \textbf{49.7} \\
\bottomrule
\end{tabular}
\caption{Performance comparison of ToG and AnchorRAG using additional LLM backbones (Claude4.5-haiku and Gemini-2.5-flash) on GrailQA datasets.}
\label{tab:additional_models}
\end{table}

\section{Case Study}
\begin{figure}[bp]
	\centering  
	\includegraphics[width=\linewidth]{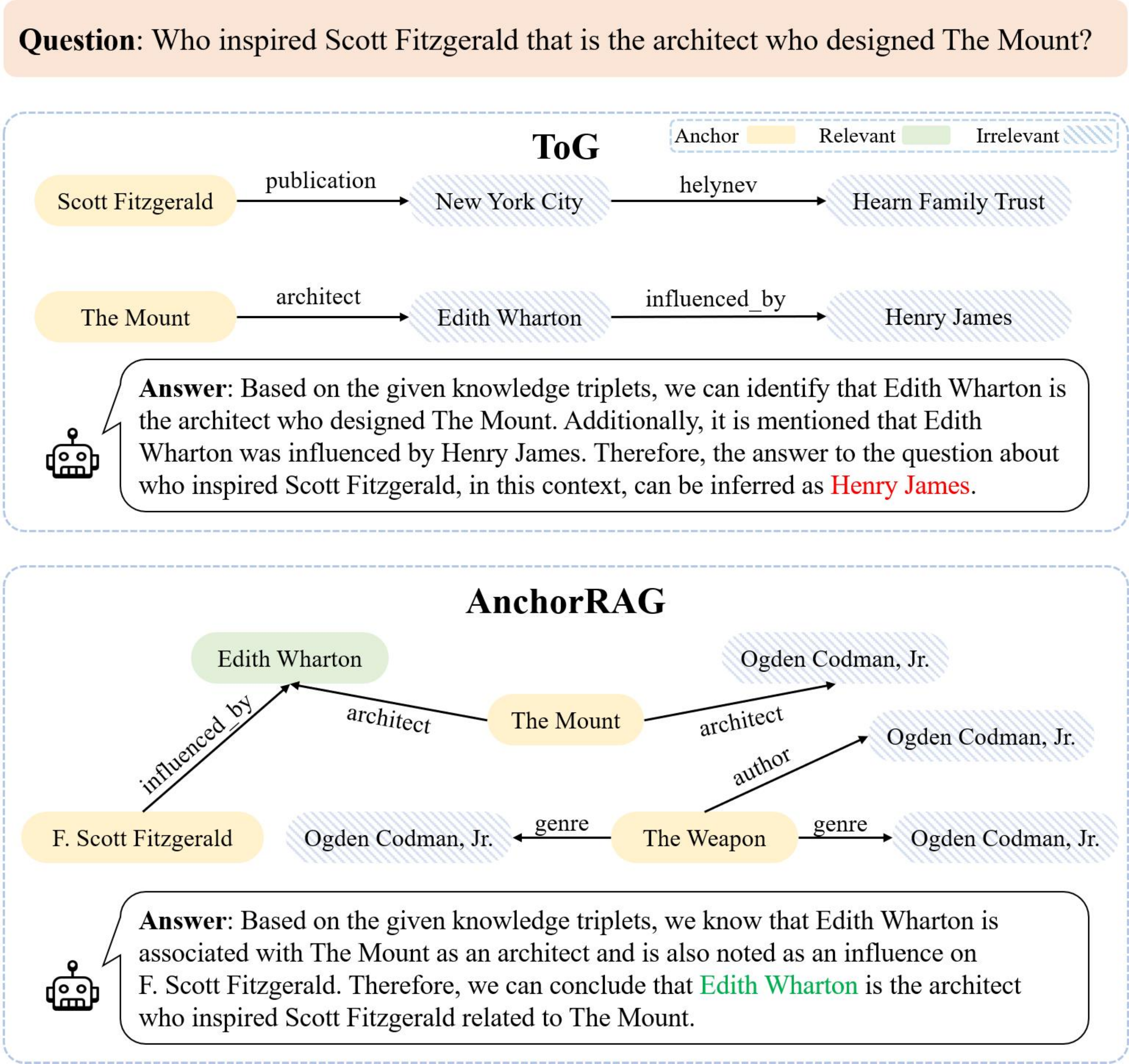}
	\caption{Comparisons of retrieval process between a baseline reasoning (ToG) and our method (AnchorRAG) on a complex question. The entities marked in yellow are anchor entities, marked in green are correct answer entities. The ones marked in blue and striped are invalid entities.}
	\label{fig:case study}
\end{figure}

To intuitively demonstrate the superiority of our method in knowledge graph retrieval, we conduct a case study that compares our approach with the ToG baseline on a representative complex question from the CWQ dataset: “Who inspired F. Scott Fitzgerald, and who was the architect that designed The Mount?”. As shown in Figure \ref{fig:case study}, we can observe that ToG can't locate both the anchor entities, resulting in meaningless reasoning paths and generating an incorrect answer. In contrast, our multi-agent framework can effectively identify the anchor entities for the given questions by the entity grounding mechanism. With assigning multi retriever agents to independently explore different reasoning paths, AnchorRAG accurately locates the correct answer entity ``Edith Wharton". Additionally, this case highlights the robustness of our method. For example, although it initially considers an incorrect entity (``The Weapon") as an anchor entity, it promptly detects the irrelevance of the following retrieval path with the rough pruning and fine filtering method, and terminates the corresponding agent, which prevents unnecessary computation overhead. 

\newpage
\onecolumn
\section{Prompts}
\label{app:prompts}
The prompts used in our method are detailed in this section.

\begin{center}
\begin{minipage}{1.0\columnwidth}
    \vspace{2mm}
        \centering
    \begin{tcolorbox}[title=Prompt for Candidate Entity Generation] 
        \small
        Extract all topic entities from the given multi-hop question. Topic entities are proper nouns, named entities, or specific concepts that are crucial for retrieving external knowledge. They may come from different sub-questions that contribute to the final answer. If there are multiple topic entities, separate them with commas.
\vspace{5pt}
    
    \texttt{In-Context Few-shot}
\vspace{5pt}
    
        Question: \texttt{<Question>}\\
    \end{tcolorbox}
    \vspace{1mm}

\end{minipage}
\end{center}

\begin{center}
\begin{minipage}{1.0\columnwidth}
    \vspace{2mm}
        \centering
    \begin{tcolorbox}[title=Prompt for Relations Pruning] 
        \small
        Please retrieve \%s relations (separated by semicolon) that contribute to the question and rate their contribution on a scale from 0 to 1 (the sum of the scores of \%s relations is 1).
\vspace{5pt}
    
    \texttt{In-Context Few-shot}
\vspace{5pt}
    
        Question: \texttt{<Question>}\\
        Topic Entity: \texttt{<Topic Entity>}\\
        Relations: \texttt{<Relations>}\\
    \end{tcolorbox}
    \vspace{1mm}

\end{minipage}
\end{center}

\begin{center}
\begin{minipage}{1.0\columnwidth}
    \vspace{2mm}
        \centering
    \begin{tcolorbox}[title=Prompt for Entities Pruning] 
        \small
        Please score the entities' contribution to the question on a scale from 0 to 1 (the sum of the scores of all entities is 1).
\vspace{5pt}
    
    \texttt{In-Context Few-shot}
\vspace{5pt}
    
        Question: \texttt{<Question>}\\
        Relation: \texttt{<Relation>}\\
        Entities: \texttt{<Entities>}\\
    \end{tcolorbox}
    \vspace{1mm}

\end{minipage}
\end{center}

\begin{center}
\begin{minipage}{1.0\columnwidth}
    \vspace{2mm}
        \centering
    \begin{tcolorbox}[title=Prompt for Triples Filtering] 
        \small
        Given the question and a list of knowledge triples, identify and return only the triples that are directly relevant to answering the question.
        Do not change the format of the triples. Do not generate any explanations or extra text. Return only the filtered triples as-is.
\vspace{5pt}
    
    \texttt{In-Context Few-shot}
\vspace{5pt}
    
        Question: \texttt{<Question>}\\
        Triples: \texttt{<Triples>}\\
    \end{tcolorbox}
    \vspace{1mm}

\end{minipage}
\end{center}

\begin{center}
\begin{minipage}{1.0\columnwidth}
    \vspace{2mm}
        \centering
    \begin{tcolorbox}[title=Prompt for Triples Evaluating] 
        \small
        Given a question and the associated retrieved knowledge graph triplets (entity, relation, entity), you are asked to answer whether it's sufficient for you to answer the question with these triplets and your knowledge (Yes or No).
\vspace{5pt}
    
    \texttt{In-Context Few-shot}
\vspace{5pt}
    
        Question: \texttt{<Question>}\\
        Triples: \texttt{<Triples>}\\
    \end{tcolorbox}
    \vspace{1mm}

\end{minipage}
\end{center}

\end{document}